\documentclass[sigconf]{acmart}

\usepackage{multirow}
\usepackage{makecell}
\usepackage{colortbl}
\AtBeginDocument{%
  }
\copyrightyear{2026}
\acmYear{2026}
\setcopyright{cc}
\setcctype{by-nc-nd}

\acmConference[CIKM '26]{Proceedings of the 35th ACM International Conference on Information and Knowledge Management}{November 7--11, 2026}{Rome, Italy.}

\acmBooktitle{Proceedings of the 35th ACM International Conference on Information and Knowledge Management (CIKM '26), November 7--11, 2026, Rome, Italy}

\acmISBN{979-8-4007-2539-5/2026/11}
\acmDOI{10.1145/3799682.3841021}

\begin{document}

\title{SADI: Spatial-Aware Dynamic Intervention for Mitigating Hallucinations in LVLMs}

\author{Yuanzhi Xu}
\orcid{0009-0008-5934-4808}
\additionalaffiliation{%
  \institution{Qilu University of Technology (Shandong Academy of Sciences)}
  \department{Key Laboratory of Computing Power Network and Information Security, Ministry of Education, Shandong Computer Science Center (National Supercomputer Center in Jinan)}
  \department{Shandong Engineering Research Center of Big Data Applied Technology, Faculty of Computer Science and Technology}
  \city{Jinan}
  \state{Shandong}
  \country{China}}
\additionalaffiliation{%
  \institution{Shandong Fundamental Research Center for Computer Science}
  \department{Shandong Provincial Key Laboratory of Industrial Network and Information System Security}
  \city{Jinan}
  \state{Shandong}
  \country{China}}
\affiliation{%
  \institution{Qilu University of Technology (Shandong Academy of Sciences)}
  \city{Jinan}
  \state{Shandong}
  \country{China}}
\email{10431250244@stu.qlu.edu.cn}

\author{Qian Gao}
\orcid{0000-0002-0239-8708}
\correspondingauthor
\additionalaffiliation{%
  \institution{Qilu University of Technology (Shandong Academy of Sciences)}
  \department{Key Laboratory of Computing Power Network and Information Security, Ministry of Education, Shandong Computer Science Center (National Supercomputer Center in Jinan)}
  \department{Shandong Engineering Research Center of Big Data Applied Technology, Faculty of Computer Science and Technology}
  \city{Jinan}
  \state{Shandong}
  \country{China}}
\additionalaffiliation{%
  \institution{Shandong Fundamental Research Center for Computer Science}
  \department{Shandong Provincial Key Laboratory of Industrial Network and Information System Security}
  \city{Jinan}
  \state{Shandong}
  \country{China}}
\affiliation{%
  \institution{Qilu University of Technology (Shandong Academy of Sciences)}
  \city{Jinan}
  \state{Shandong}
  \country{China}}
\email{gq@qlu.edu.cn}

\author{Jun Fan}
\orcid{0009-0003-8312-0939}
\affiliation{%
  \institution{China Telecom Digital Intelligence Technology Co, Ltd}
  \city{Jinan}
  \state{Shandong}
  \country{China}}
\email{fanjun.sd@chinatelecom.cn}

\author{Guohui Ding}
\orcid{0000-0001-9548-7701}
\affiliation{%
  \institution{Shenyang Aerospace University}
  \city{Shenyang}
  \state{Liaoning}
  \country{China}}
\email{dingguohui@sau.edu.cn}

\author{Zhenyu Yang}
\orcid{0009-0002-6188-5773}
\additionalaffiliation{%
  \institution{Qilu University of Technology (Shandong Academy of Sciences)}
  \department{Key Laboratory of Computing Power Network and Information Security, Ministry of Education, Shandong Computer Science Center (National Supercomputer Center in Jinan)}
  \department{Shandong Engineering Research Center of Big Data Applied Technology, Faculty of Computer Science and Technology}
  \city{Jinan}
  \state{Shandong}
  \country{China}}
\additionalaffiliation{%
  \institution{Shandong Fundamental Research Center for Computer Science}
  \department{Shandong Provincial Key Laboratory of Industrial Network and Information System Security}
  \city{Jinan}
  \state{Shandong}
  \country{China}}
\affiliation{%
  \institution{Qilu University of Technology (Shandong Academy of Sciences)}
  \city{Jinan}
  \state{Shandong}
  \country{China}}
\email{yzy@qlu.edu.cn}

\author{Sixue Lin}
\orcid{0009-0006-1781-8265}
\affiliation{%
  \institution{Qilu Institute of Technology}
  \city{Jinan}
  \state{Shandong}
  \country{China}}
\affiliation{%
  \institution{University of Nottingham Ningbo China}
  \department{School of Computer Science, Faculty of Science and Engineering}
  \city{Ningbo}
  \state{Zhejiang}
  \country{China}}
\email{scxsl5@nottingham.edu.cn}

\author{Yuteng Xiao}
\orcid{0000-0003-0773-638X}
\additionalaffiliation{%
  \institution{Qilu University of Technology (Shandong Academy of Sciences)}
  \department{Key Laboratory of Computing Power Network and Information Security, Ministry of Education, Shandong Computer Science Center (National Supercomputer Center in Jinan)}
  \department{Shandong Engineering Research Center of Big Data Applied Technology, Faculty of Computer Science and Technology}
  \city{Jinan}
  \state{Shandong}
  \country{China}}
\additionalaffiliation{%
  \institution{Shandong Fundamental Research Center for Computer Science}
  \department{Shandong Provincial Key Laboratory of Industrial Network and Information System Security}
  \city{Jinan}
  \state{Shandong}
  \country{China}}
\affiliation{%
  \institution{Qilu University of Technology (Shandong Academy of Sciences)}
  \city{Jinan}
  \state{Shandong}
  \country{China}}
\email{yutengxiao@qlu.edu.cn}

\renewcommand{\shortauthors}{Yuanzhi Xu et al.}
\begin{abstract}
Large Vision-Language Models (LVLMs) persistently suffer from object hallucinations, generating plausible yet factually incorrect outputs that fabricate objects non-existent in the visual input. Existing mitigation methods typically fall into three paradigms: fine-tuning with targeted data, which incurs prohibitive computational costs; applying contrastive decoding, which introduces high inference latency and may penalize beneficial language priors; or intervening in attention weights via heuristic hard-truncation or head-homogeneous shifts, operations that may discard information from selected attention paths. Building upon the third paradigm, we present Spatial-Aware Dynamic Intervention (SADI). SADI discards conventional rigid truncations in favor of a spatial-aware soft mask to adaptively correct visual-semantic misalignments. Specifically, it establishes a robust median consensus among multiple attention heads to anchor reliable visual representations. Guided by the spatial variance across these heads, SADI dynamically allocates intervention weights and employs a soft mask to gently suppress anomalous attention distributions. As a lightweight, training-free inference solution, our method reframes attention modulation as a smooth, continuous recalibration process. Ultimately, SADI effectively rectifies semantic drift while maintaining comparable generation fluency and description richness in our evaluations. Extensive experiments demonstrate that SADI significantly reduces both sentence- and instance-level hallucinations while maintaining description richness. Across the evaluated CHAIR, POPE, and MME benchmarks, SADI shows consistent gains in the tested settings. Code is released at \url{https://github.com/yz-8/SADI}.
\end{abstract}






\maketitle

\begin{figure*}[t]
  \centering
  \includegraphics[width=0.91\textwidth]{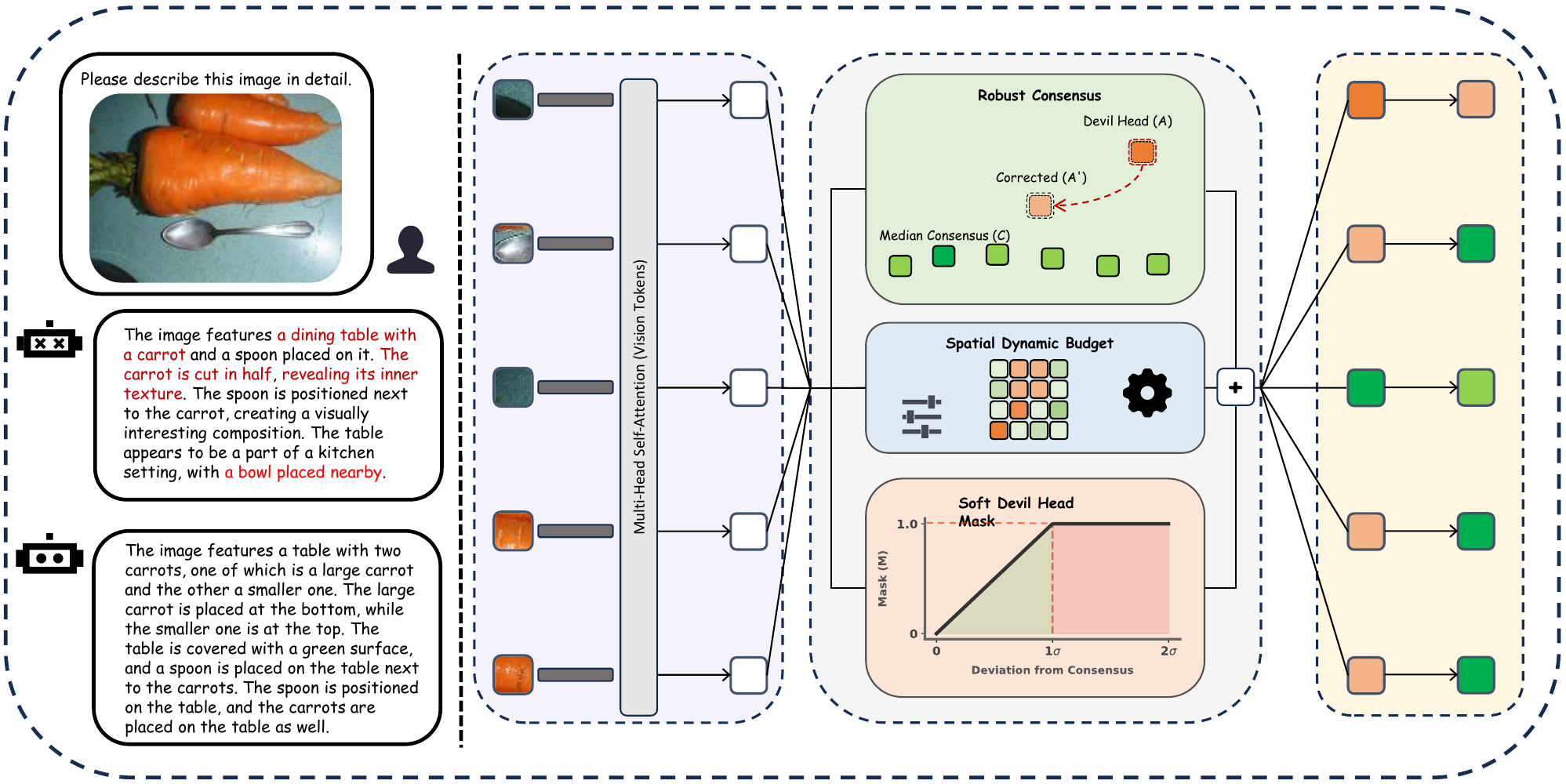} 
  \caption{Performance and pipeline of our proposed SADI. (Left) Performance comparison on hallucination mitigation. (Right) Illustration of the SADI mechanism. By leveraging robust median consensus and spatial variance-guided soft masking, SADI achieves a smooth, continuous recalibration of attention weights during a single forward pass, avoiding the severe latency and feature disruption of previous inference-time interventions.}
  \Description{A pipeline diagram illustrating the Spatial-Aware Dynamic Intervention (SADI) mechanism. The left section compares performance metrics. The right section displays a flowchart showing multi-head self-attention, robust median consensus extraction, spatial variance calculation, and adaptive soft masking applied to pre-softmax logits.}
  \label{fig:sadi_pipeline}
\end{figure*}

\section{Introduction}
Recent advancements in Large Language Models (LLMs) \cite{touvron2023llama} have revolutionized natural language processing. By integrating powerful LLM backbones with vision encoders, Large Vision-Language Models (LVLMs) \cite{zhu2024minigpt, liu2024improved, chen2023shikra, liu2023visual} map visual inputs into the textual embedding space, demonstrating unprecedented capabilities in complex multimodal tasks such as image captioning and visual question answering.

Despite these capabilities, LVLMs persistently suffer from object hallucinations \cite{li2023evaluating, rohrbach2018object, liu2024survey}. Models often generate plausible yet factually incorrect descriptions, inventing objects non-existent in the visual context. This flaw reveals a fundamental misalignment: rather than grounding generation in actual visual inputs, the model overly relies on the statistical biases and unimodal language priors inherited from the LLM \cite{sun2025understanding, li2025hidden, chen2024multi, lyu2024alleviating}. This critically limits the reliable deployment of LVLMs in high-stakes scenarios such as autonomous driving and medical diagnosis, as well as in precision-driven applications like personalized recommendation systems.

Previous efforts to mitigate object hallucinations fall into two broad categories: training-based and inference-only approaches. Training-based methods, such as fine-tuning on highly curated datasets or employing Reinforcement Learning from Human Feedback (RLHF) \cite{yu2024rlhf, liu2024mitigating, zhao2023beyond, yang2025mitigating}, effectively align model behavior but incur prohibitive computational costs and demand extensive human annotation.

To bypass the retraining overhead, inference-only strategies have been widely investigated. Contrastive decoding methods \cite{leng2024mitigating, park2025second, chen2024halc} construct positive and negative visual inputs and penalize hallucinatory tokens during generation. While effective, they necessitate multiple forward passes, which increases inference latency, and may inadvertently penalize beneficial language priors. More recently, attention-based interventions \cite{huang2024opera, jiang2025devils, sarkar2025mitigating, su2025activation} have modulated the internal mechanisms of LVLMs. Confidence-aware attention calibration \cite{fazli2026caac} and attentional vision calibration \cite{woo2025attentional} further explore training-free, soft recalibration of visual attention. However, early interventions often utilize hard-truncation or uniform head-level additions. Although recent studies explore dynamic and soft recalibrations \cite{lyu2026revealingenhancingcorevisual, sun2026mitigatingobjecthallucinationslvlms, yu2026causally}, they frequently rely on computationally intensive causal frameworks, probabilistic triggers, or mean-based aggregations that remain sensitive to activation outliers. Consequently, a trade-off persists between computational efficiency and the retention of useful intermediate representations.

In this paper, we analyze the internal attention dynamics of LVLMs during the visual information enrichment stage to address the limitations of current inference-time interventions. We propose that mitigating hallucinations is more optimally formulated as a continuous recalibration process rather than a rigid truncation of attention weights. While specific attention heads—identified as ``devils'' \cite{jiang2025devils}—disproportionately contribute to object hallucination by attending to irrelevant patches, a broader set of heads within the same layer typically maintains reliable visual grounding. The generative capability of LVLMs relies on a balance: language priors provide syntactic fluency, while multi-head self-attention encodes the visual-semantic space. Binary masking can remove selected attention paths, but it may also discard information carried by those paths and alter the balance of multi-head information aggregation.

Building upon these insights, we introduce Spatial-Aware Dynamic Intervention (SADI), a lightweight, training-free inference approach designed to seamlessly mitigate object hallucinations in LVLMs. As depicted in Figure~\ref{fig:sadi_pipeline}, departing from conventional rigid or head-homogeneous intervention strategies, SADI operates directly during the forward pass without requiring an additional decoding pass. Specifically, SADI first establishes a robust median consensus among the diverse attention heads to anchor reliable visual representations, effectively resisting the extreme outlier activations that easily corrupt standard mean-based consensus maps. Guided by the spatial variance across these heads, our method dynamically allocates an intervention budget tailored to each generated token. Instead of applying binary masking or uniform head-wide additions, SADI applies a soft mask to gently suppress the anomalous, spatially-inconsistent attention distributions. This adaptive mechanism effectively rectifies visual-semantic misalignments by enhancing the model's focus on fine-grained visual details while retaining the original attention paths and evaluating generation quality empirically, thereby providing a practical trade-off between hallucination suppression and descriptive richness.

Our main contributions are summarized as follows:
\begin{itemize}
    \item We propose Spatial-Aware Dynamic Intervention (SADI), a training-free inference method that adaptively corrects visual-semantic misalignments. By replacing hard-truncation with a median consensus mechanism and spatial variance-guided soft masking, SADI suppresses anomalous attention distributions based on fine-grained spatial contexts.
    \item We formulate inference-side attention modulation as a continuous recalibration process. This approach requires no additional decoding pass and does not require training.
\end{itemize}

Experiments on the evaluated POPE, MME, and CHAIR benchmarks with the tested LVLM families show consistent hallucination reductions without requiring external models. On CHAIR with LLaVA-1.5-7B, SADI reduces sentence-level hallucination ($C_S$) from 53.0 to 20.4 and instance-level hallucination ($C_I$) from 15.6 to 4.9, while retaining an F1 score of 76.4.

\begin{figure*}[t]
  \centering
  \includegraphics[width=0.90\textwidth]{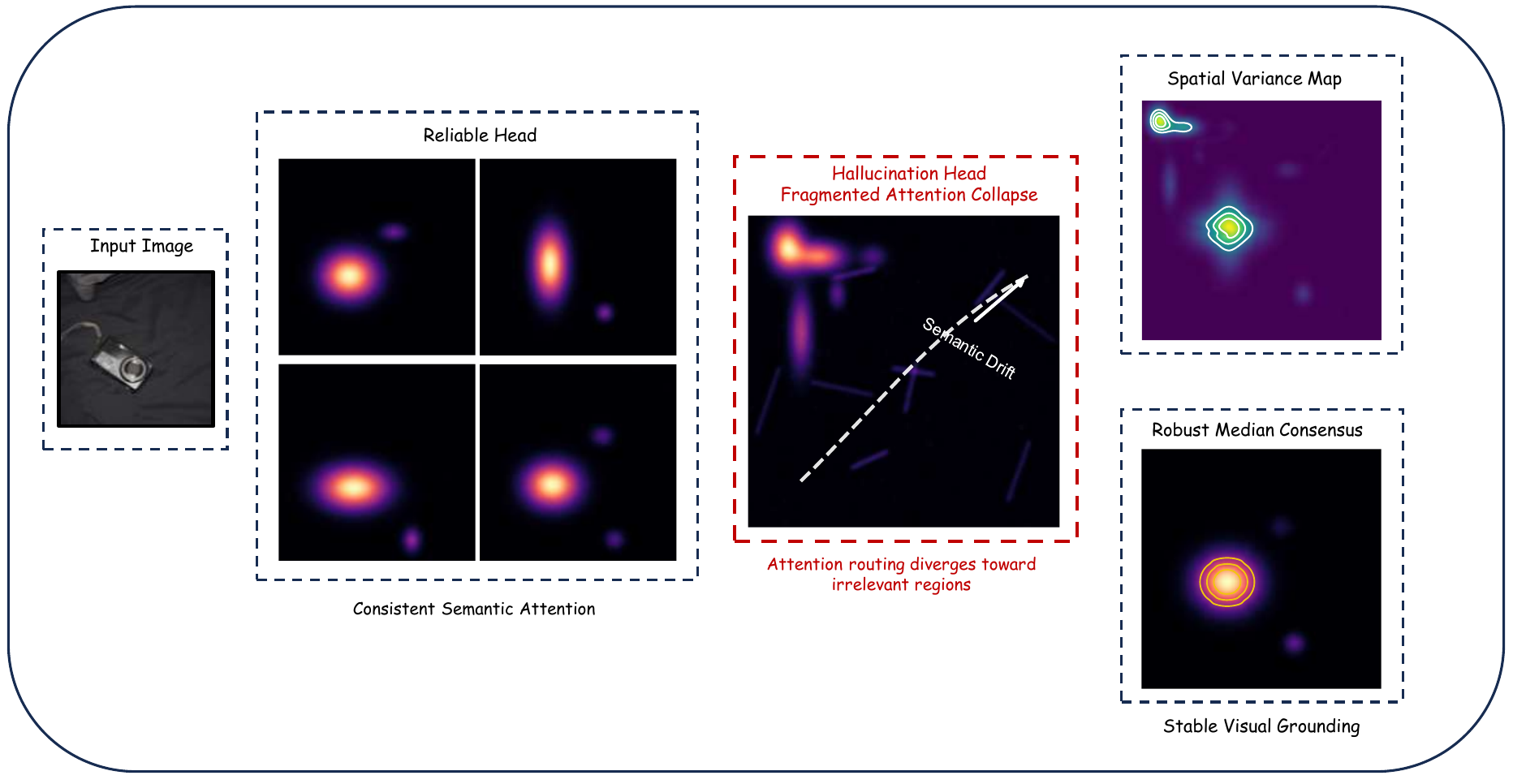} 
  \caption{Visualization of attention dynamics and the internal mechanism of SADI. While Reliable Heads exhibit consistent semantic attention grounded on the salient object, the Hallucination Head suffers from fragmented attention collapse, where attention routing severely diverges toward irrelevant background regions (Semantic Drift). To correct this, SADI extracts a Robust Median Consensus to establish stable visual grounding, and utilizes a Spatial Variance Map to dynamically guide the soft intervention budget exactly where head disagreement is highest. This is an illustrative schematic, not a quantitative before-and-after attention analysis.}
  \Description{A multi-panel visualization of attention dynamics and the internal mechanism of SADI. From left to right: the first panel shows the original input image. The second panel presents four subplots of Reliable Heads with consistent semantic attention and stable visual grounding on salient objects. The third panel illustrates a Hallucination Head with fragmented attention collapse, where a dashed arrow indicates semantic drift toward irrelevant background regions. The rightmost two panels (top to bottom) display the Spatial Variance Map and the Robust Median Consensus, respectively.}
  \label{fig:attention_visual}
\end{figure*}

\section{Related Work}

\subsection{Conventional Hallucination Mitigation in LVLMs}
Building upon the success of robust open-source Large Language Models \cite{chiang2023vicuna, touvron2023llama}, modern Large Vision-Language Models (LVLMs) \cite{zhu2024minigpt, liu2024improved, chen2023shikra, liu2023visual} demonstrate remarkable capabilities but persistently suffer from object hallucinations \cite{liu2024survey, rohrbach2018object}. To align LVLM outputs with faithful visual grounding, conventional methodologies predominantly fall into the following paradigms:

(1) \textbf{Training-Based Alignment.} Advanced techniques such as Reinforcement Learning from Human Feedback (RLHF) \cite{yu2024rlhf, xiao2025detecting}, robust instruction tuning \cite{lyu2024alleviating, liu2024mitigating}, and Direct Preference Optimization (DPO) variants \cite{zhao2023beyond, yang2025mitigating} explicitly reshape the model's behavioral distribution during the training phase. However, effectively preventing hallucinations not only necessitates robust early detection mechanisms and fine-grained benchmarking \cite{gunjal2024detecting}, but also exposes the prohibitive computational costs and the demand for large-scale, high-quality annotated datasets inherent to training, posing significant challenges for rapid model iteration.

(2) \textbf{Inference-Time Contrastive Decoding.} To bypass the prohibitive retraining overhead, training-free decoding strategies have been widely explored. Contrastive decoding methods \cite{leng2024mitigating, park2025second, chen2024halc, park2025convis} attempt to mitigate hallucinations by contrasting output logits between standard and corrupted visual inputs. Other techniques leverage mixture-of-experts \cite{liang2025mole} or image-grounded guidance \cite{zhao2024mitigating, yang2025understanding} to steer the generation trajectory at the autoregressive decoding stage.

(3) \textbf{Limitations of Conventional Paradigms.} A prominent limitation shared by both training and contrastive approaches is that they largely treat the LVLM generation process as a black box. They primarily focus on manipulating input conditions or evaluating final output logits, without actively addressing the intermediate generation dynamics. Specifically, contrastive decoding typically necessitates multiple forward passes for a single generated token, severely exacerbating inference latency. Furthermore, indiscriminate logit penalization occasionally over-corrects and destroys contextually appropriate language priors. This highlights the pressing need for white-box intervention mechanisms that can dynamically correct errors from within the internal generation pipeline.

\subsection{Internal Representation and Attention Modulation}
To overcome the black-box limitations, the paradigm most closely related to our work involves probing and intervening directly within the internal activations and attention mechanisms \cite{vaswani2017attention} of LVLMs. The evolution of this field follows a clear trajectory from passive observation to active, fine-grained intervention.

Initially, diagnostic frameworks such as Patchscopes \cite{ghandeharioun2024patchscopes}, HalLoc \cite{park2025halloc}, and investigations into the ``hidden life of tokens'' \cite{li2025hidden} successfully localized the origins of visual-semantic misalignment within specific hidden layers. Translating these diagnostic insights into active interventions, subsequent methods explored broad attention modulation. Early approaches introduced broad penalties for over-trusting certain attention patterns (e.g., OPERA \cite{huang2024opera}) or explicitly forced higher attention to visual tokens (e.g., PAI \cite{liu2024paying}).

To achieve more granular control, recent advancements have targeted specific architectural components. Studies on ``Devils in middle layers'' \cite{jiang2025devils} amplify attention scores via head-averaged activation shifts, while image-guided head suppression (SPIN) \cite{sarkar2025mitigating} explicitly identifies and masks specific hallucination-inducing attention heads. Concurrently, methods like activation steering \cite{su2025activation, chen2025ict}, null subspace projection \cite{le2412nullu}, and variational bottlenecks \cite{bai2025mitigating} apply targeted shifts to guide hidden states.

For the present study, Devils is the most direct comparison because it is an internal attention intervention on middle-layer visual-token routing and is included in our quantitative baselines. CAAC \cite{fazli2026caac} calibrates attention using confidence-aware signals, while Attentional Vision Calibration (AVC) \cite{woo2025attentional} addresses the forest-versus-trees imbalance in visual attention. These methods share SADI's training-free attention-calibration motivation, but differ in their control signals and aggregation rules. SADI instead combines an element-wise median consensus across heads with a spatial-variance-guided budget and a head-specific soft mask; Devils remains the closest baseline for the additive internal intervention studied here.

Despite these progressive refinements, a critical bottleneck persists. Early attention-head modulation predominantly relies on rigid hard-masking (e.g., binary truncation \cite{sarkar2025mitigating}) or head-insensitive global additions (e.g., \cite{jiang2025devils}), operations that may discard information carried by selected attention paths. Very recently, dynamic and soft intervention strategies have emerged \cite{lyu2026revealingenhancingcorevisual, sun2026mitigatingobjecthallucinationslvlms, yu2026causally} to address these limitations. For instance, recent studies explore activation steering in pure LLMs (also coincidentally abbreviated as SADI) \cite{wang2025semantics}, attention causal decoding \cite{tang2025seeing}, and global-local attention assemblies \cite{an2025mitigating} in LVLMs. However, unlike Wang et al. \cite{wang2025semantics}, who rely on contrastive pairs for text-semantic scaling, our proposed SADI (Spatial-Aware Dynamic Intervention) specifically utilizes multi-head median consensus to resolve visual-semantic drift. Furthermore, these advanced methods frequently depend on computationally heavy causal or probabilistic frameworks \cite{nielsen2026hallucinationanomalydynamicintervention, simhi2024constructingbenchmarksinterventionscombating}, complex external modules, or vulnerable mean-based statistical aggregations that are easily skewed by extreme outliers.

In contrast, our proposed SADI diverges from both abrupt truncations and computationally expensive dynamic frameworks. By establishing a training-free Robust Median Consensus and employing a spatial-variance-guided soft mask, SADI achieves a smooth and continuous recalibration of attention weights within a single forward computation. The measured runtime overhead and the comparison with mean-based aggregation are reported in Section~4.

\section{Method}
In this section, we present Spatial-Aware Dynamic Intervention (SADI), a training-free inference strategy designed to mitigate object hallucinations in LVLMs. SADI operates directly within the middle-to-deep transformer layers during the autoregressive decoding phase, where high-level visual semantics and language priors heavily interact (the empirical analysis of layer-wise sensitivity is detailed in Section 4). As illustrated in Figure~\ref{fig:attention_visual}, the core pipeline consists of three sequential steps: establishing a robust visual anchor via median consensus (Sec 3.2), evaluating spatial hallucination risk to allocate a dynamic intervention budget (Sec 3.3), and executing a smooth, continuous recalibration using an adaptive soft mask (Sec 3.4).

\subsection{Preliminaries: Semantic Drift}
In Large Vision-Language Models, the visual information enrichment stage relies heavily on the Multi-Head Self-Attention (MHSA) mechanism \cite{vaswani2017attention} to map visual patches to semantic tokens. Formally, during the generation of a new token, let $E_h=(E_{h,1},\ldots,E_{h,M})\in\mathbb{R}^{1\times M}$ denote the pre-softmax attention logits of head $h$ over the $M$ visual key tokens. For each head $h$ and visual position $m$, we define the scalar absolute magnitude $A_{h,m}=\lvert E_{h,m}\rvert$, and let $A_h=(A_{h,1},\ldots,A_{h,M})\in\mathbb{R}^{1\times M}$. Here, $h=1,\ldots,H$ and $m=1,\ldots,M$; all statistics below are computed independently for each visual position $m$ across the $H$ heads.

However, relying solely on this native, uncalibrated attention routing during complex multimodal reasoning is highly problematic. Our internal observations (as illustrated in Figure~\ref{fig:attention_visual}) reveal two critical pain points that precipitate and exacerbate object hallucinations:

\textbf{First}, the inherent attention mechanism is susceptible to severe Semantic Drift. While a majority of the heads maintain accurate visual grounding (Reliable Heads), a subset of anomalous heads suffers from fragmented attention collapse \cite{jiang2025devils}, assigning disproportionately high scores to irrelevant background regions. We refer to this behavior as semantic drift and quantify the corresponding cross-head disagreement using the spatial dispersion statistics defined below.

\textbf{Second}, existing mechanistic interpretability interventions sometimes use binary masks (e.g., setting selected attention paths to $0$ or $-\infty$). Such masking can remove anomalous paths, but it may also discard information carried by those paths. We therefore use a soft recalibration that retains the original paths while changing their relative logits.

These twin challenges tightly motivate our core design: rather than passively trusting the flawed native attention or destructively eliminating attention paths, we reframe hallucination mitigation as a continuous, spatial-aware recalibration process directly on the pre-softmax logits.

\subsection{Robust Median Consensus}
To smoothly correct the semantic drift, the first prerequisite is to establish a visual anchor across the attention heads. We compute the consensus independently at each visual position:
\begin{equation}
  \begin{aligned}
    C_m &= \operatorname{median}_{h=1,\ldots,H} A_{h,m},
      \quad m=1,\ldots,M,\\
    C &= (C_1,\ldots,C_M)\in\mathbb{R}^{1\times M}.
  \end{aligned}
\end{equation}
Thus, the median is an element-wise operation along the head dimension, rather than a single median over all entries. We use absolute magnitudes as an empirical statistic for the intervention; this choice does not imply that the sign of every attention logit is irrelevant. As visually demonstrated in Figure~\ref{fig:attention_visual}, the consensus vector $C$ provides a stable reference for the subsequent dynamic intervention.

\subsection{Spatial Dynamic Budget}
Having established the robust median consensus $C$, the next challenge is determining the magnitude of intervention required. Previous attention manipulation methods \cite{jiang2025devils} apply homogeneous penalties across all visual tokens, ignoring the structural complexity of the image. We hypothesize that regions with larger inter-head disagreement require stronger recalibration. For each visual position, we first define the head-wise mean:
\begin{equation}
    \mu_m = \frac{1}{H}\sum_{h=1}^{H} A_{h,m},
    \quad m=1,\ldots,M.
\end{equation}
We then compute the spatial standard deviation across heads using the unbiased estimator implemented by the attention module:
\begin{equation}
  \begin{aligned}
    S_m &= \sqrt{\frac{1}{H-1}\sum_{h=1}^{H}
      \left(A_{h,m}-\mu_m\right)^2},
      \quad m=1,\ldots,M,\\
    S &= (S_1,\ldots,S_M)\in\mathbb{R}^{1\times M}.
  \end{aligned}
\end{equation}
The standard deviation $S$ serves as a fine-grained indicator of spatial disagreement (as shown in the Spatial Variance Map in Figure~\ref{fig:attention_visual}). We perform spatial min--max normalization with $\epsilon=10^{-6}$:
\begin{equation}
    \widetilde{S}_m =
    \frac{S_m-S_{\min}}{S_{\max}-S_{\min}+\epsilon},
    \quad
    S_{\min}=\min_m S_m,\quad S_{\max}=\max_m S_m.
\end{equation}
The spatial dynamic budget is then allocated independently at each position:
\begin{equation}
    \alpha_m =
    \alpha_{\min}
    +(\alpha_{\max}-\alpha_{\min})\widetilde{S}_m,
    \quad m=1,\ldots,M.
\end{equation}
Here, $\alpha_{\min}$ and $\alpha_{\max}$ are scalar hyperparameters defining the lower and upper bounds of the intervention budget. This mechanism assigns a larger recalibration budget to positions with larger measured head disagreement.

\subsection{Adaptive Soft Masking}
With the stable consensus anchor and the spatial dynamic budget established, we introduce the core mechanism of SADI. Instead of applying rigid binary masking \cite{sarkar2025mitigating} to suppress anomalous heads, we use a soft mask to recalibrate visual--semantic misalignments. For each head $h$ and visual position $m$, we define the absolute deviation from the consensus:
\begin{equation}
    D_{h,m}=\lvert A_{h,m}-C_m\rvert.
\end{equation}
The mask is computed element-wise by normalizing this deviation with the spatial standard deviation:
\begin{equation}
    M_{h,m} =
    \operatorname{Clamp}\left(
    \frac{D_{h,m}}{S_m+\epsilon},\,0,\,1
    \right).
\end{equation}
This normalization produces a continuous mask whose value increases with the head's relative deviation from the consensus and is clipped to the interval $[0,1]$. The clipping point is an empirical design choice rather than a statistically calibrated threshold. The recalibrated pre-softmax logits are:
\begin{equation}
    \widehat{E}_{h,m}
    =E_{h,m}+\alpha_m C_m M_{h,m}.
\end{equation}
The recalibrated logits are subsequently passed through the standard softmax function. Increasing the consensus logits can reduce the relative probability assigned to other visual positions through softmax normalization. We therefore use additive-only recalibration rather than explicitly subtracting logits; preservation of representation quality is evaluated through downstream metrics and is not assumed as a formal invariance.

Because these operations are performed within the existing forward computation, SADI requires no additional decoding pass. In our runtime measurements, it incurs an approximately $1.08\times$ overhead relative to vanilla decoding.

\section{Experimental Results}

\begin{table*}[ht] 
  \centering
  \caption{Comparison of CHAIR hallucination and F1 scores across three LVLMs with max new tokens set to 512. $\Delta\%$: relative improvement over second-best ($C_S$, $C_I$); for $F_1$, relative decrease from the best method. † denotes greedy decoding.}
  \Description{A large table displaying CHAIR hallucination metrics, specifically sentence-level and instance-level hallucination rates, alongside F1 scores, for various decoding methods and models.}
  \label{tab:chair_main}
  \renewcommand\tabcolsep{9.8pt} 
  \renewcommand\arraystretch{1.3} 
  \small
  \begin{tabular}{c|rrr|rrr|rrr|rrr}
    \Xhline{1.2pt}
    \rowcolor{gray!25}
    & \multicolumn{3}{c|}{\textbf{LLaVA-1.5-7B}} & \multicolumn{3}{c|}{\textbf{LLaVA-1.5-13B}} & \multicolumn{3}{c|}{\textbf{Shikra-7B}} & \multicolumn{3}{c}{\textbf{MiniGPT-4-7B}} \\
    \cline{2-4} \cline{5-7} \cline{8-10} \cline{11-13}
    \rowcolor{gray!25} 
    \multirow{-2}{*}{\textbf{Method}} & $C_S\downarrow$ & $C_I\downarrow$ & $F_1\uparrow$ & $C_S\downarrow$ & $C_I\downarrow$ & $F_1\uparrow$ & $C_S\downarrow$ & $C_I\downarrow$ & $F_1\uparrow$ & $C_S\downarrow$ & $C_I\downarrow$ & $F_1\uparrow$ \\
    \Xhline{1pt}
    \texttt{Greedy} & 53.0 & 15.6 & 76.7 & 49.8 & 14.6 & 78.2 & 57.6 & 15.7 & 75.3 & 31.8 & 12.0 & 71.1 \\
    \texttt{Beam Search} & 55.6 & 15.4 & 77.5 & 50.4 & 13.8 & \textbf{79.0} & 59.0 & 16.3 & 74.6 & 29.2 & 9.9 & 71.2 \\
    \texttt{OPERA} \cite{huang2024opera} & 45.6 & 13.1 & \textbf{79.1} & 42.6 & 13.2 & 77.8 & 41.4 & 13.7 & 73.5 & 25.4 & 9.6 & 71.2 \\ \hline
    \rowcolor{gray!10}\texttt{† VCD} \cite{leng2024mitigating} & 58.6 & 18.2 & 72.8 & 53.6 & 15.3 & 75.8 & 56.4 & 15.5 & 75.2 & 41.4 & 14.1 & 68.2 \\
    \rowcolor{gray!10}\texttt{† PAI} \cite{liu2024paying} & 24.2 & 7.1 & 75.2 & 33.0 & 9.2 & 77.8 & 38.6 & 10.1 & \textbf{76.2} & 23.2 & 8.2 & \textbf{71.4} \\
    \rowcolor{gray!10}\texttt{† Devils} \cite{jiang2025devils} & 25.0 & 6.7 & 76.1 & 25.8 & 8.8 & 77.3 & 23.8 & 9.4 & 72.7 & 21.4 & 8.0 & 70.8 \\ \hline
    \rowcolor{gray!20}\textbf{\texttt{† SADI (Ours)}} & \textbf{20.4} & \textbf{4.9} & 76.4 & \textbf{21.2} & \textbf{5.9} & 77.5 & \textbf{20.0} & \textbf{5.9} & 74.8 & \textbf{19.2} & \textbf{5.7} & 70.4 \\
    \rowcolor{gray!20} $\Delta\%$ & +15.7 & +26.9 & -3.4 & +17.8 & +33.0 & -1.9 & +16.0 & +37.2 & -1.8 & +10.3 & +28.8 & -1.4 \\
    \Xhline{1.2pt}
  \end{tabular}
\end{table*}

\subsection{Experimental Setup}

\paragraph{Models.} To evaluate the effectiveness and generalizability of our proposed method, we conduct comprehensive experiments on three representative Large Vision-Language Models (LVLMs): LLaVA-1.5, Shikra, and MiniGPT-4. Most evaluated models are based on the 7B parameter scale to ensure a fair comparison across different architectures, with additional verification on the 13B scale to demonstrate scalability.

\paragraph{Baselines.} We compare our proposed SADI with standard decoding strategies and recent inference-time hallucination mitigation methods. Devils is the most direct quantitative comparator because it also intervenes on internal middle-layer visual attention. The baselines include: (1) Standard Decoding: Greedy decoding and Beam search decoding ($N_{beam}=5$); (2) Contrastive \& Hybrid Decoding: VCD \cite{leng2024mitigating} and PAI \cite{liu2024paying}; and (3) Attention-Guided Interventions: OPERA \cite{huang2024opera} (at decoding stage) and Devils \cite{jiang2025devils} (at internal representation stage). For fairness, all baseline methods are implemented using their official default configurations.

\paragraph{Benchmarks and Metrics.} We evaluate SADI across three diverse benchmarks to comprehensively assess both hallucination mitigation and general multimodal capabilities:
\begin{itemize}
    \item \textbf{CHAIR:} The Caption Hallucination Assessment with Image Relevance \cite{rohrbach2018object} evaluates object hallucinations in open-ended image captioning. Following previous works, we randomly select 500 images from the MSCOCO 2014 validation set \cite{lin2014microsoft}. CHAIR comprises two metrics: sentence-level ($C_S$) and instance-level ($C_I$), calculated as:
    \begin{equation}
        C_S = \frac{N_{\text{hal\_cap}}}{N_{\text{all\_cap}}}, \quad C_I = \frac{M_{\text{hal\_obj}}}{M_{\text{all\_obj}}}
    \end{equation}
    where $N_{\text{hal\_cap}}$ and $N_{\text{all\_cap}}$ denote the number of captions with hallucinated objects and all captions, while $M_{\text{hal\_obj}}$ and $M_{\text{all\_obj}}$ represent the counts of hallucinated objects and all mentioned objects, respectively. Lower values indicate fewer hallucinations. We additionally report the F1 score to monitor the accuracy and richness of the generated descriptions.
    \item \textbf{POPE:} The Polling-based Object Probing Evaluation \cite{li2023evaluating} presents a streamlined Visual Question Answering (VQA) approach to assess hallucination via binary questions (e.g., ``Is there a [object] in the image?''). It encompasses three sampling settings: Random (objects absent are chosen randomly), Popular (missing objects from a high-frequency pool), and Adversarial (co-occurring objects not present are prioritized). We report the Average F1-score across all settings to evaluate the model's resilience against statistical biases.
     \item \textbf{MME:} An extensive benchmark tailored to assess LVLMs across multiple dimensions \cite{fu2023mme}. The MME Full Set contains ten perception subtasks---\textit{Existence, Count, Position, Color, OCR, Poster, Celebrity, Scene, Landmark,} and \textit{Artwork}---and four cognition subtasks---\textit{Commonsense Reasoning, Numerical Calculation, Text Translation,} and \textit{Code Reasoning}. Following the official protocol, each image is paired with two yes/no questions. \textit{Accuracy} is the fraction of correctly answered questions, whereas \textit{accuracy+} is the fraction of images for which both paired questions are answered correctly. Their sum is the score for each subtask (maximum 200); the perception and cognition totals sum the ten and four subtask scores (maximum 2,000 and 800, respectively). This evaluation checks that our intervention does not inadvertently degrade general capabilities.
\end{itemize}

\paragraph{Implementation Details.} SADI is a training-free strategy. Based on our layer-sensitivity analysis, we restrict the intervention to intermediate layers where visual--semantic mapping is most active. The intervention layer range and the bounds of the dynamic spatial budget are configuration choices that may require adjustment for different LVLM architectures. SADI is implemented with greedy decoding and does not require an additional decoding pass; its measured runtime overhead is reported in Section~4.

\begin{table}[b] 
  \centering
  \caption{POPE evaluation results on three subsets. Average F1-score is the primary metric. $\Delta\%$: relative improvement over the second-best method.}
  \Description{A table comparing the Average F1-score of different hallucination mitigation methods on the POPE benchmark across LLaVA-1.5, Shikra, and MiniGPT-4 models.}
  \label{tab:pope}
  \renewcommand\tabcolsep{12pt} 
  \renewcommand\arraystretch{1.3} 
  \small
  \begin{tabular}{c|c|c|c}
    \Xhline{1.2pt}
    \rowcolor{gray!25}
    \textbf{Method} & \textbf{LLaVA-1.5} & \textbf{Shikra} & \textbf{MiniGPT-4} \\
    \Xhline{1pt}
    \texttt{Regular} & 79.65 & 79.80 & 69.50 \\
    \texttt{OPERA} \cite{huang2024opera} & 85.60 & 82.70 & 73.30 \\
    \rowcolor{gray!10}\texttt{VCD} \cite{leng2024mitigating} & 84.52 & 81.85 & 72.85 \\
    \rowcolor{gray!10}\texttt{PAI} \cite{liu2024paying} & 83.28 & 80.95 & 71.20 \\
    \rowcolor{gray!10}\texttt{Devils} \cite{jiang2025devils} & 86.12 & 83.45 & 74.60 \\ \hline
    \rowcolor{gray!20}\textbf{\texttt{SADI (Ours)}} & \textbf{86.73} & \textbf{84.15} & \textbf{75.50} \\
    \rowcolor{gray!20} $\Delta\%$ & +0.7 & +0.8 & +1.2
    \\
    \Xhline{1.2pt}
  \end{tabular}
\end{table}

\subsection{Main Results}

\paragraph{Results on CHAIR} As presented in Table~\ref{tab:chair_main}, SADI significantly surpasses all baselines in reducing both sentence-level ($C_S$) and instance-level ($C_I$) hallucinations. Specifically, on LLaVA-1.5-7B, SADI drastically reduces $C_S$ to 20.4 and $C_I$ to 4.9. Mechanistically, contrastive approaches like VCD inadvertently exacerbate $C_S$ to 58.6 because their global noise-injection corrupts fine-grained visual cues and delicate prompt-following priors, leading to fragmented text. Furthermore, compared to recent internal interventions like PAI and Devils, SADI achieves deeper suppression while uniquely preserving description richness (F1 score of 76.4). This is because Devils applies a rigid, head-homogeneous attention shift based on vulnerable mean aggregations, which indiscriminately distorts the native distributions of reliable heads. In contrast, SADI utilizes robust median consensus and spatial-aware soft masking to surgically target only anomalous semantic drift without disrupting the overall feature context.

\paragraph{Results on POPE} Table~\ref{tab:pope} summarizes the POPE results under random, popular, and adversarial settings. The generally lower scores in the adversarial setting are consistent with models frequently falling into the trap of statistical co-occurrence priors (e.g., hallucinating a ``mouse'' simply because a ``keyboard'' is present). Under these rigorous settings, SADI achieves an Average F1-score of 86.73 on LLaVA-1.5, the highest score among the methods compared in this table. While OPERA (85.60) attempts to penalize over-confident n-gram patterns during late-stage decoding as a symptom suppressor, SADI intervenes precisely at the internal cross-modal fusion zone. This result is consistent with the intended effect of aligning attention updates with the visual consensus, although the current evaluation does not isolate the causal contribution of this mechanism.

\begin{figure*}[t]
  \centering
  \includegraphics[width=0.91\textwidth]{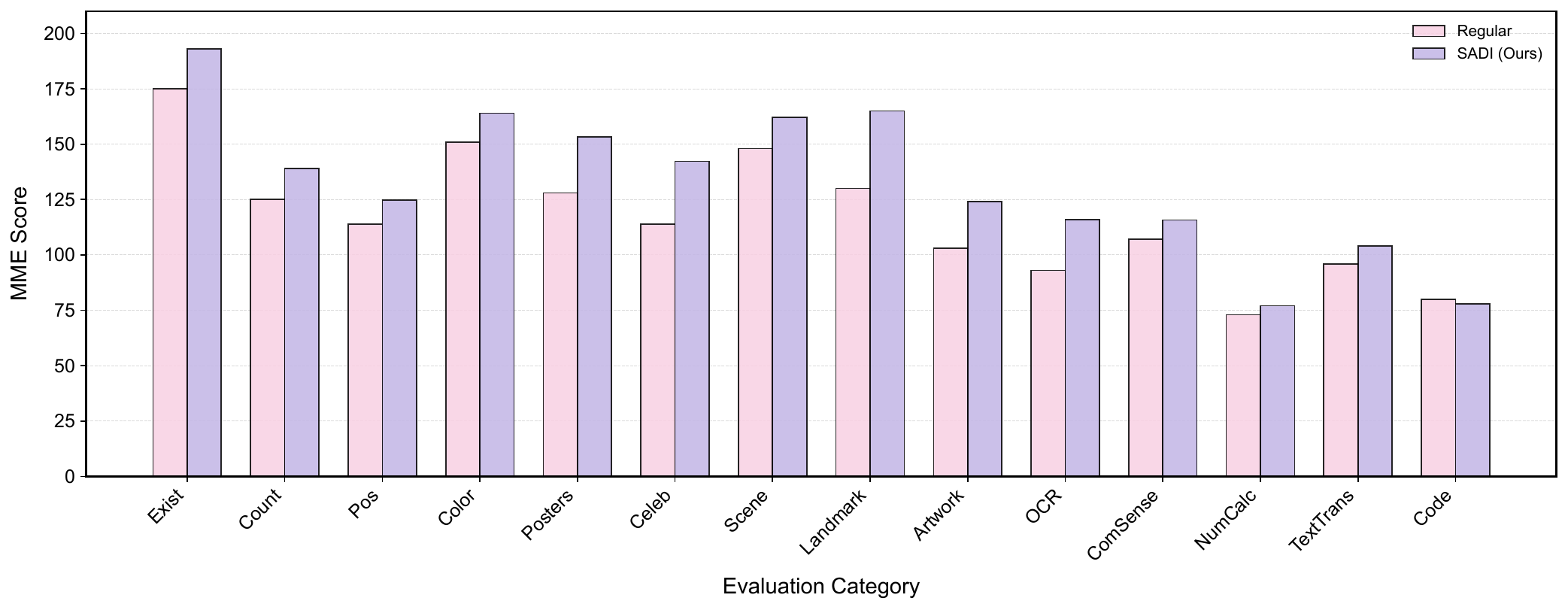} 
  \caption{Comprehensive evaluation on the MME Full Set. We compare our proposed SADI against the regular baseline across $14$ sub-tasks. SADI achieves a consistent performance boost in perception-based categories (e.g., \textit{Existence}, \textit{Count}, and \textit{Position}) and \textit{Common Sense} while firmly preserving the model's native cognitive competencies like \textit{Code Reasoning}.}
  \Description{A bar chart comparing the MME scores of the Regular baseline and SADI (Ours) across fourteen evaluation categories, including Exist, Count, Pos, Color, Posters, Celeb, Scene, Landmark, Artwork, OCR, ComSense, NumCalc, TextTrans, and Code.}
  \label{fig:mme}
\end{figure*}

\paragraph{Results on MME} Results on the MME Full Set (Figure~\ref{fig:mme}) confirm that our hallucination mitigation does not compromise general capabilities. SADI yields a consistent performance boost in fine-grained perception tasks like \textit{Existence}, \textit{Count}, and \textit{Position}. This pattern is consistent with the intended role of the spatial disagreement map, although the causal contribution of this mechanism is not isolated by the current experiments. Crucially, SADI largely maintains the model's native cognitive competencies (e.g., \textit{Code Reasoning}). Unlike methods that explicitly remove attention paths, SADI uses additive-only recalibration. The observed MME scores indicate that the evaluated cognitive capabilities are largely retained.

\subsection{Ablations and Analysis}

\paragraph{Layer Sensitivity Analysis: Locating the Semantic Drift.} To characterize the effect of intervention depth, we divide the 32 transformer layers of LLaVA-1.5 into three stages---Early (layers 0--4), Middle (layers 5--18), and Deep (layers 19--31)---and evaluate each stage on CHAIR. As detailed in Table~\ref{tab:layer}, the mitigation effect is layer-dependent. Intervening in the deep layers (19--31) yields only marginal improvements ($C_S$ drops from 53.0 to 51.6), whereas applying SADI to the early layers (0--4) reduces hallucinations but lowers F1 to 75.2. The strongest performance in this experiment is obtained in the intermediate interval (5--18), where $C_S=20.4$, $C_I=4.9$, and F1 is 76.4. These results identify the intermediate layers as a useful intervention region for the evaluated LLaVA-1.5 setting. Similar intermediate-layer behavior is observed for the other evaluated models, although the appropriate absolute indices may vary with architecture depth.

\begin{figure*}[t]
  \centering
  \includegraphics[width=0.91\textwidth]{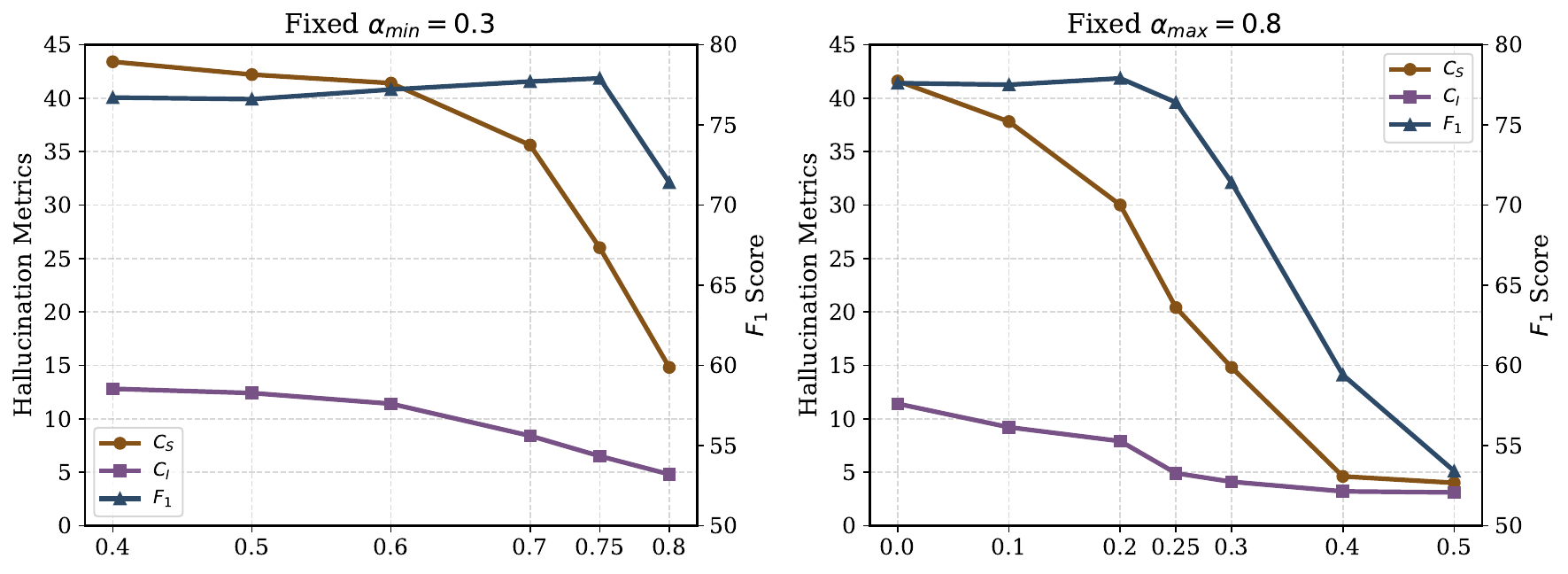} 
  \caption{Hyperparameter sensitivity analysis. (Left) Varying the upper bound $\alpha_{\max}$ with a fixed $\alpha_{\min}=0.3$. (Right) Varying the lower bound $\alpha_{\min}$ with a fixed $\alpha_{\max}=0.8$.}
  \Description{A two-panel hyperparameter sensitivity analysis plot. The left panel shows the impact of varying the upper bound of the dynamic budget, while the right panel shows the impact of varying the lower bound.}
  \label{fig:hyperparam}
\end{figure*}

\begin{table}[ht] 
  \centering
  \caption{CHAIR hallucination evaluation and F1 scores across three intervention intervals (Early: 0--4, Middle: 5--18, Deep: 19--31) on LLaVA-1.5-7B under 512 maximum generation tokens.}
  \Description{A table showing the effects of applying the SADI intervention at different transformer layer intervals on the CHAIR hallucination metrics.}
  \label{tab:layer}
  \renewcommand\tabcolsep{16pt} 
  \renewcommand\arraystretch{1.3} 
  \small
  \begin{tabular}{c|c|c|c}
    \Xhline{1.2pt}
    \rowcolor{gray!25}
    \textbf{Intervened Layers} & \textbf{$C_S \downarrow$} & \textbf{$C_I \downarrow$} & \textbf{F1 $\uparrow$} \\
    \Xhline{1pt}
    \texttt{None} & 53.0 & 15.6 & \textbf{76.7} \\
    \rowcolor{gray!10}\texttt{Layers 19--31} & 51.6 & 13.8 & 76.1 \\
    \rowcolor{gray!10}\texttt{Layers 0--4} & 24.6 & 9.0 & 75.2 \\ \hline
    \rowcolor{gray!20}\textbf{\texttt{Layers 5--18}} & \textbf{20.4} & \textbf{4.9} & 76.4 \\
    \Xhline{1.2pt}
  \end{tabular}
\end{table}

\paragraph{Adaptive Soft Masking vs. Static Mean Addition.} To validate the advanced design of SADI, we specifically compare it with the intervention mechanism proposed in Devils \cite{jiang2025devils}. As formulated in their methodology, the Devils approach attempts to find a ``faithful direction'' by averaging the absolute attention scores across all heads (a mean-based consensus) and adding it homogeneously across all attention heads for each visual token with a fixed hyperparameter $\alpha$. However, this strategy suffers from two critical flaws: (1) the mean operation is highly susceptible to extreme outliers (severe semantic drift in anomalous heads), and (2) indiscriminately applying the same head-averaged boost to all heads can distort the native attention distributions of reliable heads. In contrast, SADI resolves these bottlenecks through two design changes. By substituting the vulnerable mean with a Robust Median Consensus, SADI effectively isolates extreme outliers. Furthermore, instead of a global static addition, SADI deploys a Spatial-Aware Dynamic Budget combined with an Adaptive Soft Mask. This ensures that the recalibration is only applied to highly disputed spatial regions and weights heads according to their deviations from the consensus. As reflected in the CHAIR results (Table~\ref{tab:chair_main}, reducing $C_S$ from 25.0 in Devils to 20.4 in SADI), this fine-grained, adaptive recalibration significantly outperforms rigid, mean-based global addition while preserving high descriptive richness.

\paragraph{Robustness to Generation Length.}
In long-form open-ended generation, LVLMs are highly prone to a ``snowballing'' effect \cite{zhong2024investigating}: as the generated text lengthens, the model increasingly attends to its own prior textual outputs rather than the visual input, causing initial minor semantic drifts to compound into severe object hallucinations. To evaluate SADI's robustness against this phenomenon, we assess the hallucination metrics on the CHAIR benchmark across varying maximum generation lengths ($\{64, 128, 256\}$). As detailed in Table~\ref{tab:chair_tokens}, standard decoding strategies suffer from drastic performance degradation as generation length increases. For instance, the sentence-level hallucination ($C_S$) of Greedy decoding surges from $23.8$ at 64 tokens to $53.0$ at 256 tokens. In stark contrast, SADI consistently achieves the lowest $C_S$ and $C_I$ across all length settings. More importantly, SADI demonstrates exceptional stability: even when the maximum token limit is extended to 256, its $C_S$ remains tightly controlled at $20.8$. While achieving this profound hallucination suppression necessitates a slight trade-off in descriptive richness (yielding an $F_1$ of $76.7$, perfectly matching the Greedy baseline) compared to methods like OPERA ($F_1$ of $79.1$), OPERA suffers from a massively inflated hallucination rate ($C_S$ of $45.6$, more than double that of SADI). This explicit trade-off strongly validates that SADI prioritizes faithful visual grounding over unconstrained linguistic extrapolation, effectively preventing the accumulation of semantic drift over long generation trajectories.

\begin{table}[ht] 
  \centering
  \caption{CHAIR evaluation on LLaVA-1.5-7B across different max new tokens $\{64, 128, 256\}$. $\dagger$: greedy decoding.}
  \Description{A table comparing CHAIR hallucination metrics across different maximum generation token lengths, demonstrating SADI's robustness to long-form generation.}
  \label{tab:chair_tokens}
  \renewcommand\tabcolsep{2.5pt} 
  \renewcommand\arraystretch{1.3}
  \small
  \begin{tabular}{c|rrr|rrr|rrr}
    \Xhline{1.2pt}
    \rowcolor{gray!25}
    & \multicolumn{3}{c|}{\textbf{max 64}} & \multicolumn{3}{c|}{\textbf{max 128}} & \multicolumn{3}{c}{\textbf{max 256}} \\
    \cline{2-4} \cline{5-7} \cline{8-10}
    \rowcolor{gray!25} 
    \multirow{-2}{*}{\textbf{Method}} & $C_S\downarrow$ & $C_I\downarrow$ & $F_1\uparrow$ & $C_S\downarrow$ & $C_I\downarrow$ & $F_1\uparrow$ & $C_S\downarrow$ & $C_I\downarrow$ & $F_1\uparrow$ \\
    \Xhline{1pt}
    \texttt{Greedy} & 23.8 & 8.0 & 74.7 & 52.4 & 15.5 & 76.7 & 53.0 & 15.6 & 76.7 \\
    \texttt{Beam} & 17.6 & 6.0 & \textbf{75.1} & 54.4 & 14.8 & 77.6 & 55.6 & 15.4 & 77.5 \\
    \texttt{OPERA} & 19.0 & 6.3 & 74.4 & 44.2 & 12.9 & \textbf{78.8} & 45.6 & 13.1 & \textbf{79.1} \\ \hline
    \rowcolor{gray!10}\texttt{† VCD} & 26.0 & 9.3 & 73.3 & 56.8 & 16.9 & 74.5 & 58.6 & 18.2 & 72.8 \\
    \rowcolor{gray!10}\texttt{† PAI} & 19.8 & 6.2 & 74.0 & 24.2 & 7.3 & 75.2 & 24.2 & 7.2 & 75.2 \\
    \rowcolor{gray!10}\texttt{† Devils} & 18.2 & 5.4 & 73.9 & 24.8 & 7.1 & 76.0 & 25.0 & 7.1 & 76.1 \\ \hline
    \rowcolor{gray!20}\textbf{\texttt{† SADI (Ours)}} & \textbf{15.4} & \textbf{4.7} & 74.1 & \textbf{20.4} & \textbf{5.4} & 76.5 & \textbf{20.8} & \textbf{5.3} & 76.7 \\
    \Xhline{1.2pt}
  \end{tabular}
\end{table}

\begin{table}[ht] 
  \centering
  \caption{Inference time comparison (ms) among five decoding methods on an NVIDIA H800 GPU across three generation length settings (20/50/80 tokens). Relative slowdown over the LLaVA-v1.5 baseline is shown in parentheses.}
  \Description{A table comparing the inference time in milliseconds for different decoding methods across varying generation lengths on an NVIDIA H800 GPU.}
  \label{tab:efficiency}
  \renewcommand\tabcolsep{6.0pt} 
  \renewcommand\arraystretch{1.3} 
  \small
  \begin{tabular}{c|c|c|c}
    \Xhline{1.2pt}
    \rowcolor{gray!25}
    \textbf{Method} & \textbf{20-Token Len} & \textbf{50-Token Len} & \textbf{80-Token Len} \\
    \Xhline{1pt}
    \texttt{LLaVA-v1.5} & 405 ($\uparrow\times1.0$) & 935 ($\uparrow\times1.0$) & 1440 ($\uparrow\times1.0$) \\
    \rowcolor{gray!10}\texttt{OPERA}~\cite{huang2024opera} & 1372 ($\uparrow\times3.4$) & 3294 ($\uparrow\times3.5$) & 5717 ($\uparrow\times4.0$) \\
    \rowcolor{gray!10}\texttt{VCD}~\cite{leng2024mitigating} & 989 ($\uparrow\times2.4$) & 2032 ($\uparrow\times2.2$) & 3078 ($\uparrow\times2.1$) \\
    \rowcolor{gray!10}\texttt{Devils}~\cite{jiang2025devils} & 422 ($\uparrow\times1.04$) & 972 ($\uparrow\times1.04$) & 1498 ($\uparrow\times1.04$) \\ \hline
    \rowcolor{gray!20}\textbf{\texttt{SADI (Ours)}} & 438 ($\uparrow\times1.08$) & 1009 ($\uparrow\times1.08$) & 1555 ($\uparrow\times1.08$) \\
    \Xhline{1.2pt}
  \end{tabular}
\end{table}

\begin{figure*}[t]
  \centering
  \includegraphics[width=0.88\textwidth]{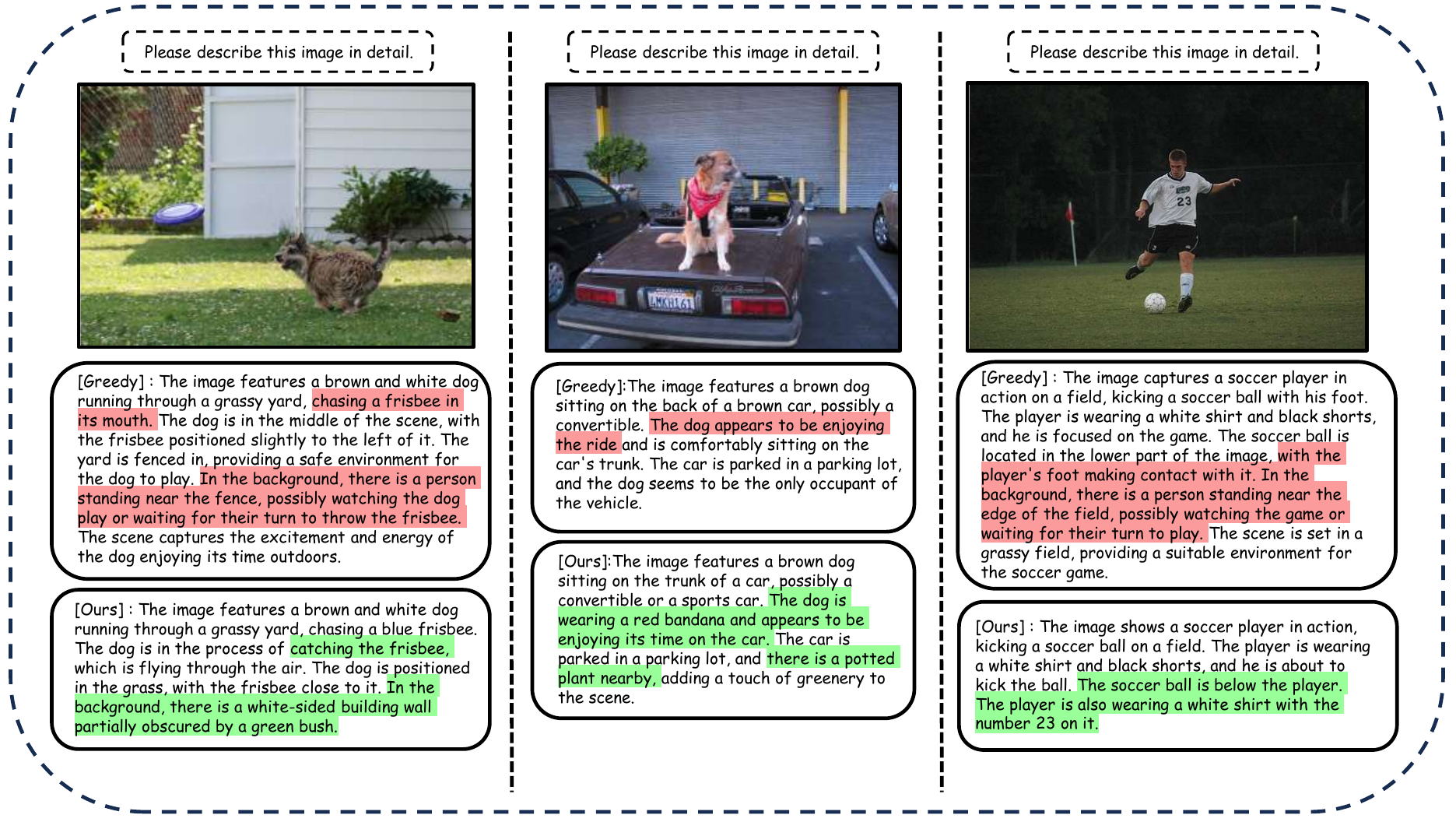} 
  \caption{Qualitative comparison between Greedy decoding and SADI on LLaVA-1.5. Red text highlights hallucinations from the baseline, while green text denotes corrected factual descriptions. The panel compares only these two decoding settings.}
  \Description{A qualitative analysis chart showcasing three case studies of hallucination mitigation by SADI compared to the Greedy baseline. Each case features an input image on top, followed by a red-highlighted description from the Greedy baseline containing hallucinations (e.g., false persons, false actions, missed attributes), and a green-highlighted description from SADI providing faithful and detailed descriptions.}
  \label{fig:qualitative_cases}
\end{figure*}

\paragraph{Inference Efficiency Analysis.} A core advantage of SADI is its ability to mitigate object hallucinations without incurring prohibitive inference latency. To validate its practical deployment efficiency, we evaluate the decoding time of SADI against the vanilla LLaVA-1.5 and other representative inference-time mitigation baselines (VCD, OPERA, and Devils). The evaluations are conducted on a single NVIDIA H800 GPU, recording the time required to generate sequences of 20, 50, and 80 tokens. As detailed in Table~\ref{tab:efficiency}, contrastive decoding methods like VCD require multiple forward passes for a single token, resulting in a significant computational bottleneck (exhibiting over a $2.1\times$ slowdown). Similarly, OPERA's heavy reliance on complex beam-search and retrospection strategies incurs a severe latency penalty, slowing down the inference by up to $4.0\times$. In contrast, SADI operates directly on the pre-softmax logits using highly parallelized basic tensor operations (e.g., median and standard deviation) strictly within a single forward pass. Consequently, SADI maintains a decoding speed nearly identical to the vanilla LLaVA-1.5 model, introducing a small but nonzero computational overhead of approximately $1.08\times$. While Devils exhibits a marginally lower latency ($1.04\times$), it suffers from inferior hallucination mitigation performance compared to SADI. These measurements support the practical deployment of SADI in the evaluated setting.

\paragraph{Hyperparameter Sensitivity Analysis.} 
To evaluate the robustness of our dynamic intervention budget, we conducted a sensitivity analysis on the essential hyperparameters $\alpha_{\min}$ and $\alpha_{\max}$ (dynamic-budget equations above). As shown in Figure~\ref{fig:hyperparam}, varying these bounds reveals a clear performance trade-off. While escalating intervention strength continuously suppresses hallucinations (both $C_S$ and $C_I$ decrease), exceeding a specific threshold triggers an over-correction cliff, indiscriminately suppressing beneficial semantic priors and severely degrading description richness ($F_1$ score drops significantly). Consequently, empirically anchoring the parameters strictly at the edge of this cliff (e.g., $\alpha_{min} \approx 0.25, \alpha_{max} \approx 0.80$) provides a favorable trade-off in the evaluated setting, with substantial hallucination reduction and comparable language quality. The results suggest a favorable region in this experiment, while the optimal settings may vary across architectures.

\paragraph{Qualitative Case Analysis.}
To visually verify how SADI translates into explicit output calibration, we provide side-by-side case studies across prominent visual scenarios in Figure~\ref{fig:qualitative_cases}. The baseline model (Greedy decoding) heavily exhibits semantic drift, projecting unimodal language priors to fabricate missing entities (e.g., inventing ``a person near the fence'' on both the lawn and soccer field) or over-interpreting static scenes into phantom actions (e.g., claiming a parked car is ``enjoying the ride''). Conversely, SADI cleanly dampens these hallucination trajectories by enforcing robust median grounding, ensuring actions are bound tightly to physical states (e.g., ``about to kick the ball''). Crucially, as highlighted in green, SADI meticulously captures fine-grained, localized visual details overlooked by the baseline—such as the dog's ``red bandana'' and the jersey ``number 23''—empirically substantiating our claims of refining semantic alignment without sacrificing descriptive richness.

\paragraph{The Danger of Explicit Subtraction.} 
SADI employs an additive-only recalibration on pre-softmax logits. To validate this, we test an ``Add-Subtract'' variant that explicitly penalizes anomalous background regions ($- \beta \odot \text{BackgroundMask} \odot M_h$). As Table~\ref{tab:ablation_subtract} shows, explicit subtraction counter-intuitively exacerbates hallucinations. As the penalty $\beta$ increases, $C_S$ surges from 20.4 to 41.4. This degradation suggests that explicit subtraction can remove contextual signal. Un-salient background tokens may still provide useful spatial context, so forcefully reducing them can increase reliance on language priors. Consequently, it generates highly fluent (evidenced by rising F1 scores) but visually ungrounded text. SADI instead amplifies consensus regions and relies on softmax normalization to reduce their relative alternatives; representation quality is assessed empirically.

\begin{table}[ht] 
  \centering
  \caption{Ablation on intervention mechanisms (Additive vs. Add-Subtract) on LLaVA-1.5.}
  \Description{A table presenting an ablation study comparing the CHAIR metrics of the proposed Additive-Only mechanism against an Add-Subtract variant with different penalty values.}
  \label{tab:ablation_subtract}
  \renewcommand\tabcolsep{13pt} 
  \renewcommand\arraystretch{1.3}
  \small
  \begin{tabular}{c|c|c|c}
    \Xhline{1.2pt}
    \rowcolor{gray!25}
    \textbf{Intervention Type} & \textbf{$C_S \downarrow$} & \textbf{$C_I \downarrow$} & \textbf{F1 $\uparrow$} \\
    \Xhline{1pt}
    \rowcolor{gray!5}\texttt{Greedy} & 53.0 & 15.6 & 76.7 \\ \hline
    \rowcolor{gray!10}\texttt{Add-Subtract ($\beta=0.5$)} & 31.6 & 7.2 & 77.8 \\ 
    \rowcolor{gray!10}\texttt{Add-Subtract ($\beta=1.0$)} & 36.2 & 9.5 & 78.2 \\ 
    \rowcolor{gray!10}\texttt{Add-Subtract ($\beta=2.0$)} & 41.4 & 10.9 & \textbf{78.4} \\ \hline
    \rowcolor{gray!20}\textbf{\texttt{Additive-Only (SADI)}} & \textbf{20.4} & \textbf{4.9} & 76.4 \\
    \Xhline{1.2pt}
  \end{tabular}
\end{table}

\section{Conclusion}
In this paper, we introduced Spatial-Aware Dynamic Intervention (SADI), a lightweight, training-free inference strategy to alleviate object hallucinations in Large Vision-Language Models (LVLMs). We identified that during the cross-modal fusion stage, some existing inference-time interventions use hard truncation or rigid masking, which can discard information from selected attention paths. To overcome this, SADI reframes hallucination mitigation as a smooth, continuous recalibration process. By establishing a robust median consensus among attention heads and deploying a spatial variance-guided dynamic budget, our method applies an adaptive soft mask to gently suppress anomalous semantic drift. Extensive experiments demonstrate that SADI effectively grounds the model's generation in visual realities, significantly reducing both sentence- and instance-level hallucinations on the CHAIR and POPE benchmarks, while effectively maintaining the model's native general capabilities on MME. SADI achieves these results without an additional decoding pass and incurs a measured runtime overhead of approximately $1.08\times$ in our evaluation.

\section{Limitations}
SADI has several limitations that bound the scope of its claims. First, the median consensus assumes that a majority of attention heads provide a reliable visual anchor. In crowded, occluded, or multi-object scenes, a minority head may instead capture a useful secondary object, so the consensus and soft mask can over-suppress information that should be retained. Second, the intervention layer range and the bounds of $\alpha$ are selected empirically and may require architecture-specific adjustment. Third, lower hallucination scores do not by themselves completely rule out a conservative-generation effect; although the reported F1, POPE, and MME results provide complementary checks, they do not fully disentangle reduced hallucination from fewer risky object mentions. Finally, Figure~\ref{fig:attention_visual} is an illustrative mechanism schematic rather than a quantitative validation of before-and-after attention changes, and the current evaluation does not include mask-level alignment or density/occlusion-stratified analysis.

\section{Future Work}
Building upon the remarkable efficacy of SADI, several promising avenues remain for future exploration to further elevate LVLM capabilities. First, while our robust median consensus effectively anchors salient objects, future work will explore dynamic clustering-based consensus mechanisms to intelligently preserve minority attention heads that correctly capture secondary details in highly complex, multi-object scenes, thereby preventing potential over-correction. Second, we aim to develop an auto-adaptive framework that dynamically computes the optimal intervention layers and spatial budget boundaries on the fly, enabling truly zero-shot, hyperparameter-free deployment across diverse LVLM architectures. Finally, we plan to leverage SADI's continuous spatial variance signals as dense rewards within rigorous alignment training paradigms (e.g., RLHF or DPO), seamlessly bridging our inference-time intervention with fundamental parametric refinement.

\begin{acks}
This work was funded by the projects ZR2022MF333 and ZR2024QF053 supported by Shandong Provincial Natural Science Foundation, in part by the Pilot Project for Integrated Innovation of Science, Education, and Industry of Qilu University of Technology (Shandong Academy of Sciences) 2026ZDCX01, and in part by the project of the Key Laboratory of Computing Power Network and Information Security, Ministry of Education under Grant No. 2023ZD028
\end{acks}

\section*{Generative AI Usage Statement}
The generative AI tools used in this paper are confined to language-level assistance, including grammar checking, automatic spelling correction, wording refinement, and clarity improvement. They were not employed to generate any original content or research ideas. All substantive content has been authored and verified solely by the authors.

\bibliographystyle{ACM-Reference-Format}
\bibliography{main}

\end{document}